\documentclass{article}

\usepackage{arxiv}

\usepackage[utf8]{inputenc} 
\usepackage[T1]{fontenc}    
\usepackage{hyperref}       
\usepackage{url}            
\usepackage{booktabs}       
\usepackage{amsfonts}       
\usepackage{nicefrac}       
\usepackage{microtype}      
\usepackage{lipsum}
\usepackage{graphicx}
\graphicspath{ {./images/} }
\usepackage{graphicx}
\usepackage{textcomp}

\usepackage{tikz}
\usepackage{comment}
\usepackage{amsmath,amssymb} 
\usepackage{color}

\usepackage[accsupp]{axessibility}  
\usepackage{textcomp}

\title{Inverted Semantic-Index for Image Retrieval}

\author{
 Ying Wang \\
 School of Information Science and Engineering\\
 Southeast University\\
 Nanjing, China \\
  \texttt{wying@seu.edu.cn} \\
}

\begin{document}
\maketitle

\begin{abstract}
This paper addresses the construction of inverted index for large-scale image retrieval. The inverted index proposed by J. Sivic brings a significant acceleration by reducing distance computations with only a small fraction of the database. The state-of-the-art inverted indices aim to build finer partitions that produce a concise and accurate candidate list. However, partitioning in these frameworks is generally achieved by unsupervised clustering methods which ignore the semantic information of images. In this paper, we replace the clustering method with image classification, during the construction of codebook. We then propose a merging and splitting method to solve the problem that the number of partitions is unchangeable in the inverted semantic-index. Next, we combine our semantic-index with the product quantization (PQ) so as to alleviate the accuracy loss caused by PQ compression. Finally, we evaluate our model on large-scale image retrieval benchmarks. Experiment results demonstrate that our model can significantly improve the retrieval accuracy by generating high-quality candidate lists. 

\end{abstract}

\section{Introduction}

Inverted indices are commonly used for image retrieval~\cite{ivf}~\cite{imi}~\cite{PQawared}. Firstly, a large dataset of images are extracted into a set of feature descriptors. Then a typical inverted index is built according to a codebook consisting of codewords, i.e. a set of centroids constructed by clustering. The points belonging to each codeword are stored in the corresponding list. The purpose of an inverted index is then to select several relevant codewords and generate a candidate list of vectors that lie close to any query vector. The application of inverted index avoids the exhaustive distance computations. Thus the retrieval efficiency is much higher with the help of an index structure.

The state-of-the-art inverted indices~\cite{imi}~\cite{g+p}~\cite{PQawared} attempt to build a finer segmentation on the database in order to provide a more precise searching scope. The inverted multi-index (IMI)~\cite{imi} decomposes the feature space into two orthogonal subspaces and clusters each subspace independently. The grouping and pruning method with inverted index (IVF+G+P)~\cite{g+p} computes the subcentroids under each partition and generates a much finer codebook based on the subcentroids. However, these indices still construct the partitions according to the centroids calculated by unsupervised clustering methods such as K-means. The basement on clustering regardless of semantic information may cause three problems. Firstly, these indexes are not able to be combined with local features~\cite{delf}, since the dimension of local features is unfixed and the comparison approach is usually based on a geometric verification RANSAC. Secondly, the number of reclaimed partitions cannot be determined without the prior knowledge of the data distribution(Fig.~\ref{fig2}).  Thirdly, the accuracy drops considerably~\cite{imi}
~\cite{PQawared} when the cluster-based indexes are combined with Approximate Nearest Neighbor Search (ANN Search) approaches~\cite{pq}~\cite{hnsw}~\cite{nvtree}. The most essential reason of their disadvantages is the excessive dependence on features and data distribution.

\begin{figure}[t]
\includegraphics[width=\textwidth]{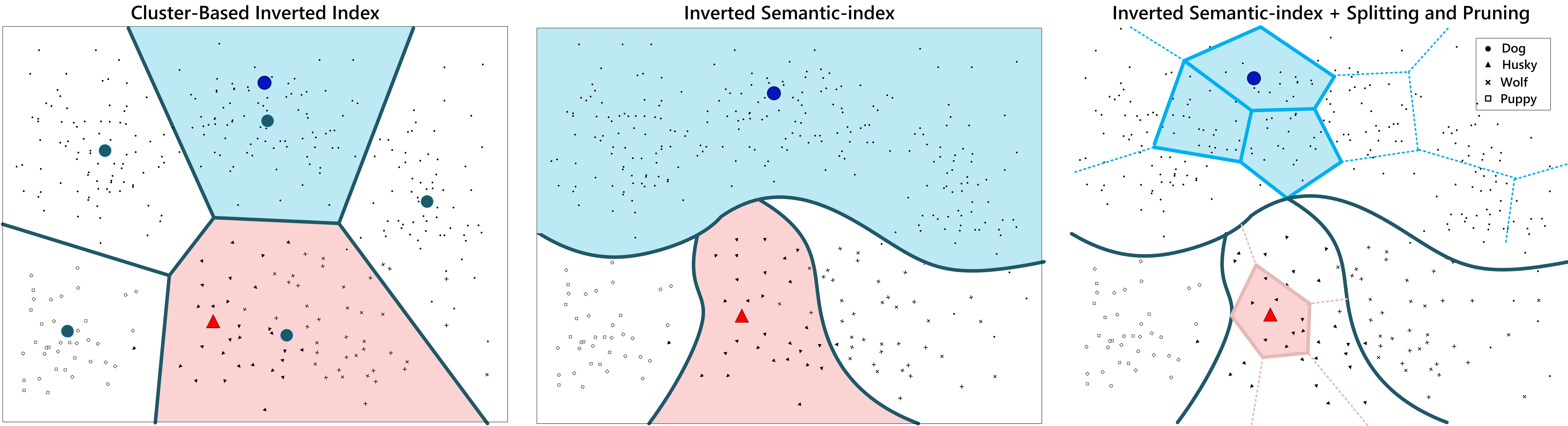}
\caption{The higher DEF $\alpha$ or QEF $\beta$ is ,the searching scope is bigger. Accompanying this expansion is the increase in accuracy and the decrease in acceleration efficiency.} \label{fig3}
\end{figure}

The goal of this paper is to introduce and evaluate a new index structure for instance image retrieval called the {\it inverted semantic-index.}  This index divides the database based on labels predicted by classifiers instead of centroids produced by clustering. The advantage of semantic-index is its ability to be combined with both local feature and global feature(Tab.~\ref{tab1}), thanks to the fact that partition boundaries generated by semantic-index are independent of features. Besides, the number of the reclaimed partitions in semantic-index is normalized to 5 (Fig.~\ref{fig2}). Owing to the outstanding performance of image classifiers, the arrangement of image data can adapt to the data distribution robustly. For example, when the amount of semantic-related images in the database is large for a query image, the relevant 5 partitions will include more points. On the contrary, the categories rare to the database own less images. In addition, the  quality of the candidate list generated by semantic-index is much higher thanks to the participation of semantic information. Consequently, semantic-index results in more accurate retrieval regardless of the size of the database and it can even mitigate the accuracy loss for PQ (Tab.~\ref{tab3}). All in all, the successful data arrangement based on semantic information can also be extended to other large-scale searches, such as video retrieval and face recognition.

The contributions of this paper can be summarized as three aspects:
\begin{enumerate}
\renewcommand{\labelenumi}{(\theenumi)}
    \item We propose an inverted semantic-index for large-scale instance image retrieval, which can be combined with both local feature and global feature. With the semantic-index, the suitable number of reclaimed cells can be normalized to 5. In a nutshell, our index has data-driven adaptability. 
    \item We propose a label merging method as well as a partition splitting method to solve the problem that the size of the codebook is unchangeable in semantic-index.
    \item  The experiment results prove that with higher-quality candidate lists, our inverted semantic index can achieve an excellent accuracy performance on all-scale  image retrieval benchmarks and alleviate the precision degradation brought by the approximation of PQ. 
\end{enumerate}

\section{Related Work}
\subsubsection{Image Feature Extraction} is the approach of representing images as mathematical vectors. There are two categories of feature extractors, i.e. local feature~\cite{delf} and global feature~\cite{Gem}. The local feature is usually based on key point detection and description~\cite{delf}, while global feature is commonly based on the global information extracted from the layers of CNNs~\cite{alexnet}~\cite{vggnet}~\cite{resnet}~\cite{Gem}. Unlike the global feature, the unfixed dimension and geometric comparison of local feature bring the unfitness with the large scale cluster-based index. Many works such as VLAD~\cite{vlad}, Fisher Vector~\cite{fvtree} and vocabulary tree~\cite{VOC} focus on aggregation method of local feature, in which the lower accuracy decline is pursued. In our experiments, we apply a local feature algorithm DELF~\cite{delf} and a global feature extractor Gem~\cite{Gem} to transform the real world image datasets to feature vectors.  

\subsubsection{Product quantization(PQ)} After the generation of candidate list by inverted indices, the Approximate Nearest Neighbor Search (ANN Search) approaches~\cite{pq}~\cite{hnsw}~\cite{nvtree} are used to re-rank the candidates efficiently. The Product Quantization~\cite{pq} has been recognized as the most successful ANN method. It is a compression method which aims to represent high-dimensional vectors in binary codewords with ignorable loss~\cite{pq}. PQ divides a vector $x\in R^D$ into $M$ sub-vectors $x=[x_1,\dots,x_M],x_m \in R^{\frac{D}{M}}$. Hence the feature space is divided into M orthogonal subspaces. The codebooks $\{R_1,\dots,R_M\} $ are built on these subspaces, where each sub-vector is clustered into K-bit codewords $\{r_1^m,\dots,r_{2^K}^m\},r_i^m\in R^{\frac{D}{M}}$. It is proved that the M-byte code is an efficient approximation of $x$~\cite{pq}. Then the computation of Euclidean distances efficiently is performed via the ADC procedure using lookup tables~\cite{pq}:
\begin{equation}
||q-x||^2\approx||q-[r_{i_1}^1,\dots,r_{i_M}^M]||^2=\sum_{m=1}^M||q_m-r
_{i_m}^m||^2,
\end{equation}
in which the distance from query $m$-th sub-vector $q_m$ to the $2^K$ codewords $r_{1,\dots,2^K}^m$ of the $m$-th codebook $R_m$ can be precomputed and stored in lookup tables. Owing to the outstanding compression performance and computational speed, PQ has been recognized as the most popular and successful solution and inspires many works in computer vision and machine learning~\cite{imi}~\cite{g+p}~\cite{PQawared}.

\subsubsection{Image Classification} The image classifiers~\cite{alexnet}~\cite{vggnet}~\cite{resnet} based on supervised CNNs achieve excellent performance in recent years. For instance, the top-$5$ accuracy of ResNet~\cite{resnet} is up to $94.75\%$. This indicates a phenomenon that the images with the same content are much likely to own common elements in the top-5 labels. Therefore, we propose two assumptions based on which we construct the semantic-index based. The first assumption is that the top-5 labels of two similar images may overlap. As a result, if we store the images into semantic partitions, reclaiming 5 partitions relevant to the query image is enough to recall all the target images. The second assumption is that even if the two similar images belong to unseen labels which are not in these $1k$ labels, there must be overlaps on the top-5 wrong labels(Fig.~\ref{fig4}). Thus no matter what is the semantic contribution of the database, reclaiming 5 out of $1k$ is still effective. The application of labels to inverted index also appears in the work~\cite{semantic}. Nevertheless, it aims at the content-based retrieval, while we concentrate more on the instance retrieval.

\section{Our Approach}

\subsection{The Inverted Semantic-index}
\begin{figure}[t]
\includegraphics[width=\textwidth]{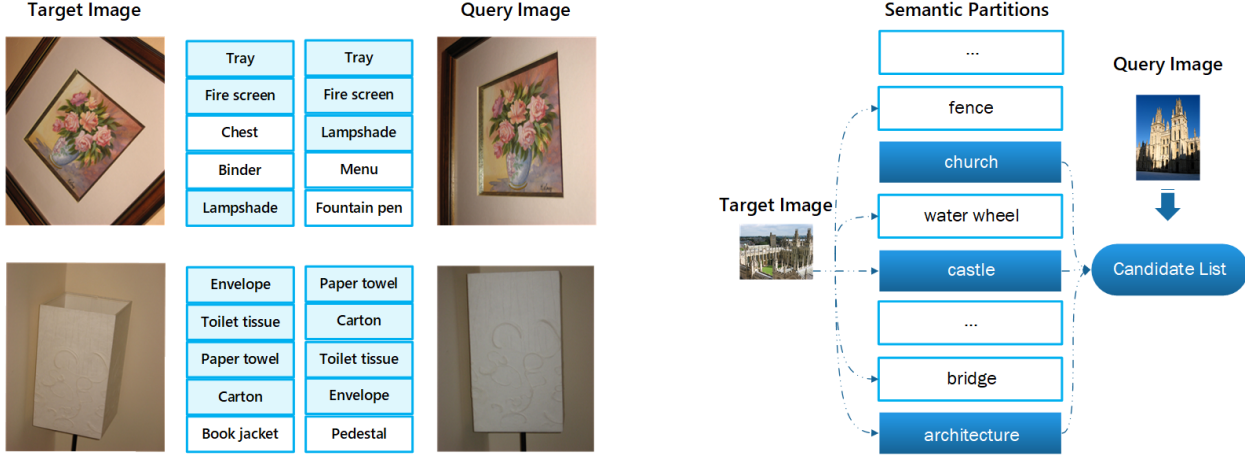}
\caption{The images on the left from Holidays dataset is classified by ResNet. It is obvious that the top-5 labels of query and target own common elements in spite of that none of the labels is the real semantic caption `picture frame' and `lampshade'. Thus the second assumption about unseen labels can be corroborated, which is proposed in the related work section. On the right hand, the framework about the contribution of the Inverted Semantic-index and selection of searched cells is displayed. The fusion parameter $\alpha$ is 5 and expanding parameter $\beta$ is 3 here.} \label{fig4}
\end{figure}
Now we introduce how an inverted semantic-index is organized which is displayed in the Fig.~\ref{fig4}. Unlike the cluster-based indexes, the codebook of semantic-index $W=\{l_1,\dots,l_i,\dots,l_{N_l}\}$ is built based on the categories~\cite{imagenet} rather than centroids, where $N_l$ is the number of labels. Then the possibility $p_i$ that the database image $X$ belongs to the $i$-th category is predicted by ResNet~\cite{resnet}, where $p_i(X)$ satisfies $\sum_{i=1}^{N_l} p_i(X)=1$. All these confidence values are stored in a vector:
\begin{equation}
p(X)=[p_1(X),\dots,p_i(X),\dots,p_{N_l}(X)].
\end{equation}
Then the top-$\alpha$ labels of image $X$ can be determined by ranking the confidences computed above, which processing is represented as the function $T_{\alpha}$. Next, each image in database is recorded into the lists related to the top-$\alpha$ labels. Thus, the database is divided into $1k$ partitions in which the $i$-th partition $W_i$ can be represented as a set: 
\begin{equation}
W_i^{\alpha}=\{x|i\in \text{T}_{\alpha} \left( p(X) \right) \},
\end{equation}
where x denotes the feature extracted from image $X$ and the variable $\alpha$ here is named as fusion parameter. Inferred from the property that the Top-5 accuracy of ResNet is up to 94.75\%~\cite{resnet}, the fusion parameter $\alpha$ is generally set to 5 so as to make sure the images are recorded into the right label cell. At this point, the construction of the inverted semantic-index is completed. 

Now we address the problem that which partitions are selected to generate the candidate list in the retrieval procedure. Unlike cluster-based indexes where distances are compared to centroids to reclaim cells, the inverted semantic-index still relies on classifiers for partition selection. For each query image $Q$, the top-$\beta$ labels are confirmed via the function $T_{\beta}$ mentioned above. Then select the partitions associated with these $\beta$ labels and reclaim the belonging points to generate the candidate list:

\begin{equation}
L^{\alpha}_{\beta}=\{W_j^{\alpha}|j\in \text{T}_{\beta}\left( p(Q) \right)\},
\end{equation}
where $\beta$ is called expanding parameter. For the same reason as $\alpha$, $\beta$ is typically set to 5 to ensure the correct-label partition is searched.

All in all, the details of generating candidate list via the semantic-index are described above, where the fusion parameter $\alpha$ controls the overlap degree of the partitions and the expanding parameter $\beta$ decides the size of the candidate list. The improvement of $\alpha$ and $\beta$  will bring about an increase in accuracy and a decrease in speed. On the contrary, the decrease of $\alpha$ and $\beta$ will reduce the accuracy and searching time.

\subsection{Merging and Splitting of Partitions}
The candidate list can be produced out of the database via the inverted semantic-index construction method explained in the last section. However, there is a problem brought by the application of semantic labels: the number of labels is fixed and unchangeable in our semantic-index. Thus the size of the codebook is unable to be adjusted according to the scale of database. Therefore, a method of merging and splitting the partitions is in demand to solve this problem. Further more, the method of merging similar labels is more and more important when the classifiers produce more labels nowadays~\cite{labelgraph1}~\cite{labelgraph2}~\cite{semantic}.

\begin{figure}
\includegraphics[width=\textwidth]{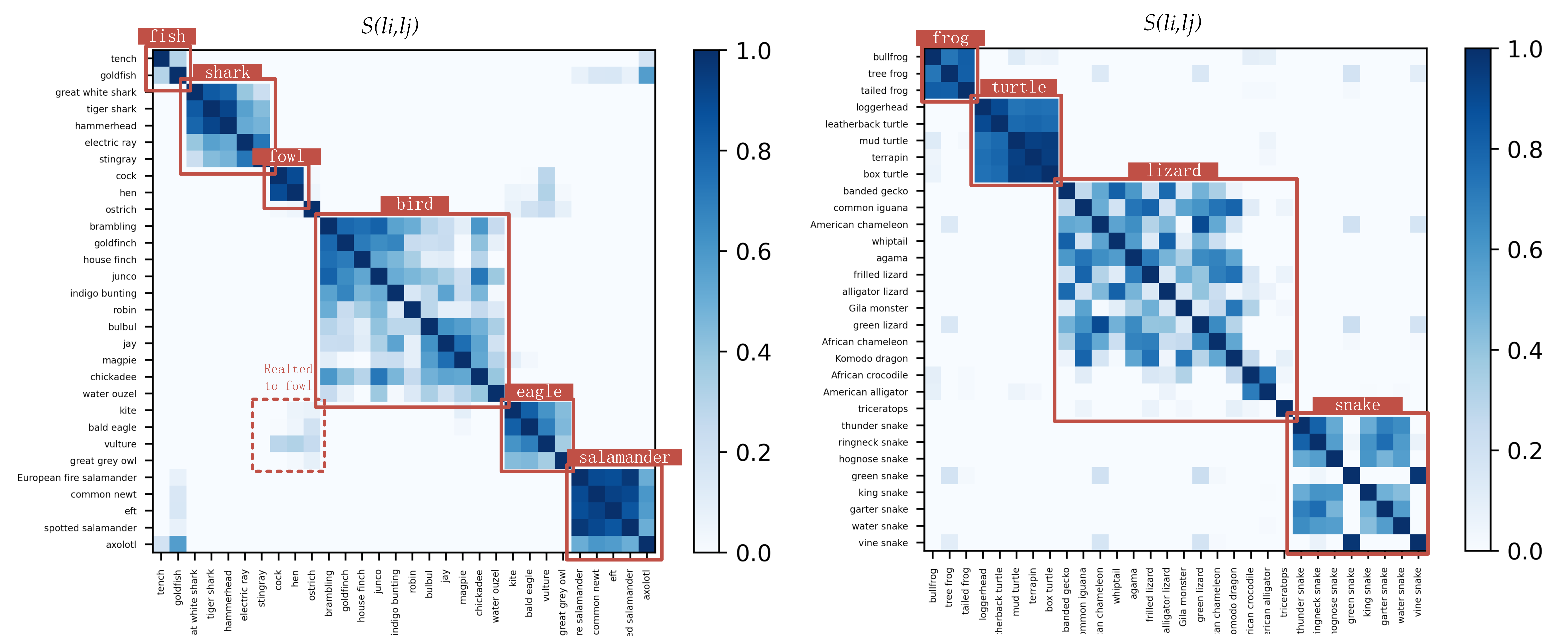}
\caption{Part of the correlation coefficients between labels calculated by Eq.~(\ref{equ7}) are shown here. The child classes of each big category are gathered in low-rank blocks. And the method can catch the semantic relation between `fowl' and `eagle' as well.} \label{fig5}
\end{figure}

Let us introduce the label merging method first. It is well known that there are gaps between labels without the similarity quantitative method~\cite{labelgraph1}~\cite{labelgraph2}. Thus how to measure the similarity of two semantic label is an interesting question. Based on the phenomenon that the similar or related labels often co-occur in the top-5 predictions, we propose a statistical label merging method: for each image in ILSVRC dataset~\cite{imagenet}, we count the times $t_{i,j}$ that label $l_i$ and label $l_j$ appear in the top-5 at the same time and record them into the matrix $C\in R^{N_l\times N_l}$:

\begin{equation}
C=\left[
\begin{matrix}
 t_{1,1}      & t_{1,2}      & \cdots & t_{1,N_l}      \\
 t_{2,1}      & t_{2,2}      & \cdots & t_{2,N_l}      \\
 \vdots & \vdots & \ddots & \vdots \\
 t_{N_l,1}      & t_{N_l,2}      & \cdots & t_{N_l,N_l}      \\
\end{matrix}
\right].
\end{equation}
 Since the similar or related labels have similar distributions on the co-occurrence times with all other labels, the $i$-th row of the matix $C_i$ is a semantic distribution featue of the $i$-th label:
\begin{equation}
C_i=[t_{i,1},\dots,t_{i,N_l}].
\end{equation}
Then we calculate the correlation coefficient $r$ of $C_i$ and $C_j$ and build the similarity evaluation function between label $l_i$ and label $l_j$ based on it:

\begin{equation}
s(l_i,l_j)=r(C_i,C_j)=\frac{Cov(C_i,C_j)}{\sigma_{i}\sigma_{j}},
\label{equ7}
\end{equation}
where $Cov(C_i,C_j)$ is the covariance of $C_i$ and $C_j$, meanwhile $\sigma_i$ and $\sigma_j$  mean the standard deviations of $C_i$ and $C_j$. Given the similarity quantitative method, we can merge the partitions and construct the multi-scale semantic index structure with clustering methods, such as K-means or Ratio Cut. Part of the label similarity results are shown in Fig.~\ref{fig5}.

Next, the partition splitting method is explained here. When the database is large, a finer division is in demand. Thus the splitting method of semantic partitions is required to locate a accurate and concise scope. Inspired by the brilliant idea of pruning in IVF+G+P~\cite{g+p}, we split each semantic partition into $L=10$ parts by K-means, hence the codebook size is enlarged to $10k$. When we reclaim one cell from ten under each semantic partition, the pruning parameter $\tau$ is set to $0.1$. The performance of merging and splitting will be shown in Tab.~\ref{tab4} and Tab.~\ref{tab3} by experiments.

\subsection{ANN Search with Inverted Semantic-index}

When the large-scale database requires extortionate memory size, PQ is applied to compress the data into binary code. For instance, the IVF-ADC system~\cite{pq} uses $x\approx c+[r_1,\dots,r_M]$ to approximate the dataset point, while the Inverted Multi-index (IMI)~\cite{imi} system uses $x\approx[c_1,c_2]+[r_1,\dots,r_M]$ to compress each data. In these cluster-based cases, the centroids are not only the approximating participator of the point cluster but also the rules of the partition boundaries. Nevertheless, with the application of classifier on inverted semantic-index, the nonlinear boundary is produced by the convolution neural network (CNN)~\cite{resnet}. As a result, we do not have the central points to participate in the approximation computation. In order to combine our inverted semantic-index with PQ, the centroids $c^i\in R^D$ of each semantic partition $W_i$ are precomputed: 

\begin{equation}
c^i=\text{argmin}_c \sum_{j=1}^{N}{\begin{Vmatrix}x_j-c\end{Vmatrix}^2} , c\in W_i=\{x_1,\dots,x_j,\dots,x_N\}.
\end{equation}
Following the product quantization and Multi-D-ADC ideas~\cite{imi}, the feature $x$ can be approximated by a sum:

\begin{equation}
x \approx c^i+[r_1,\dots,r_M],
\end{equation}
where the vector $[r_1,\dots,r_M]$ is a compressed result calculated by PQ. Thus, the distance from the query to the point $x$ in the database is approximated by:
 
 \begin{equation}
 \begin{aligned}
\begin{Vmatrix}q-x\end{Vmatrix}^2 \approx 
\begin{Vmatrix}q\end{Vmatrix}^2-2\left \langle q,c^i\right \rangle-2\sum_{m=1}^M{\left \langle q_m,r_m\right\rangle}+\sum_{m=1}^M{\begin{Vmatrix}c^i_m+r_m\end{Vmatrix}^2},
\end{aligned}
\label{equ2}
\end{equation}
while the cosine similarity between the query and the point in partitions is represented as:
 
 \begin{equation}
 \begin{aligned}
s(q,x)\approx \frac{\left \langle q,c^i\right \rangle+{\sum_{m=1}^M{\left \langle q_m,r_m\right\rangle}}} {\begin{Vmatrix}q\end{Vmatrix}\cdot \sqrt{\sum_{m=1}^m{\begin{Vmatrix}c^i_m+r_m\end{Vmatrix}}^2}}.
\end{aligned}
\label{equ1}
\end{equation}
Typically, we can precompute the dot-products $\left \langle q,c^i\right \rangle$ and $\left \langle q_m,r_m\right \rangle$ and store the results in lookup tables~\cite{imi}. Therefore, these terms can be reused during the retrieval procedure. Further more, the term $\begin{Vmatrix}c_n^i+r_n\end{Vmatrix}^2$ is independent of query $q$. The precomputation and storage can also be carried out by lookup tables.

\section{Experiments}
In this section we evaluate our inverted semantic-index on image retrieval benchmarks. The inverted semantic-index is realized by Python on the Intel(R) Xeon(R) Gold 6240 CPU @ 2.60GHz in a single thread mode, in spite of that the implement of Gem~\cite{Gem} is based on GPU which only influences the preparation procedure of converting images to feature vectors. In the experiments, we used the implementation of IVF, PQ and IMI from FAISS library~\cite{faiss} for comparison.

For comparison, the codebook size $K$ is set to $N_l=1k$ in the state-of-the-art cluster-based index such as IVF and IVF+G+P~\cite{g+p}. As for the grouping parameter $L$ and pruning parameter $\tau$ in IVF+G+P are set to 64 and 0.5 as mentioned~\cite{g+p}. For the reason that the implement of multi-index in Faiss Library~\cite{faiss} can not set $K$ to $1k$, we set the codebook size $K$ of each orthogonal subspace to $2^{10}$ so as to approximate $1k$. In addition, all the PQ compression~\cite{pq} are completed with the parameter $M=8$ and $K=8$. In other words, all the feature vectors are compressed to 8 bytes when the indexes are combined with ADC comparison. In experiments, the retrieval performance is evaluated by $mAP$ and the quality of candidate list is evaluated by recall, where the parameter $R@{1,10,100}$ means the percentage of recalled target images when the length of candidate list is $1, 10$ and $100$.

\subsection{Datasets}
Most of prior inverted indices evaluate their indexes on the SIFT-1M~\cite{pq}, Deep-1B~\cite{deep-1b} and GIST-1M~\cite{pq} benchmarks which are consist of low-level descriptors without the source images. However, our semantic-index is aimed at image retrieval and the raw image data is indispensable for image classification, thus we evaluate the performance of our index on the real world instance image retrieval benchmarks~\cite{oxford5k}~\cite{paris6k}~\cite{holidays} rather than the ANN datasets~\cite{pq}~\cite{deep-1b}. The image retrieval benchmarks are shown bellow:
\subsubsection{Oxford5k}~\cite{oxford5k} contains 5062 building images captured in Oxford with 55 query images. Each query has 6 to 221 target images. The differences between target images and query images include perspective conversion, light change, occlusion, etc. These are the common challenges for instance image retrieval. 
\subsubsection{Paris6k}~\cite{paris6k} is consist of 6412 architecture images from Paris, which also has 55 query images. The ground-truth amount of each query counts 51 to 289. The challenges are similar to the Oxford5k dataset.
\subsubsection{Holidays}~\cite{holidays} has 1491 personal holiday pictures with 500 query images. Each query has 2 to 13 target images. The retrieval difficulty is relatively low. However its content are more varied, which can benefit verifying the validity of the semantic-index under unseen labels and prove our second assumption proposed in the image classification part of the related work section. 
\subsubsection{Flickr100k and Flickr1M}~\cite{oxford5k} are distractors generally combined with the datasets mentioned above, which contain 100 thousand and 1 million images from Flickr respectively. The semantic distributions of these datasets are more widespread.

Next we apply two state-of-the-art feature extraction algorithms which are designed for image retrieval so as to transform the images to descriptors. One is the local feature DELF~\cite{delf} and the other is global feature Gem~\cite{Gem}. 
\subsection{The Determination of Fusion and expanding parameters}
\begin{figure}
\includegraphics[width=\textwidth]{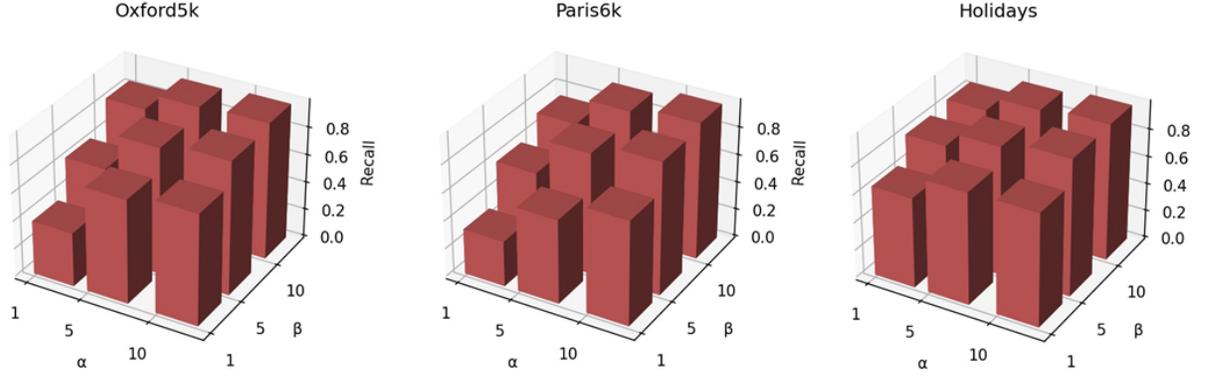}
\caption{The increase of $\alpha$ or $\beta$ can broaden the size of candidate list.} \label{fig1}
\end{figure}

In the section 3.1, we infer based on prior knowledge that the target images may be all recalled when fusion parameter $\alpha$ and expanding parameter $\beta$ are 5. Now we test the correctness of this assumption in this section by experiments. As shown in Fig.~\ref{fig1}, the recall increases with $\alpha$ and $\beta$. When the $\alpha$ and $\beta$ are both set to $10$, the recall on Oxford5k, Paris6k and Holidays reach to 97.88\%, 97.71\% and 98.53\%, which indicate that almost all the target images are reclaimed. Meanwhile the percentage of searching scope comparing with the whole database is 59.70\%, 71.00\% and 32.25\%  on these three datasets  (The scope compression efficiency is higher in Holidays for the reason that its semantic distribution is more dispersed, while the samples in Oxford5k and Paris6k mostly gather in the partitions related to `building').
When $\alpha$ and $\beta$ are set to 5, the recall is still high at 90.45\%, 86.61\% and 92.69\%  respectively, meanwhile the range ratio is lower at 45.19\%, 43.32\% and 16.44\% compared with top10-10. In brief, with a larger candidate list at top10-10, the recall is increased and the searching scope is larger. In order to get a trade-off between the accuracy and speed, we perform the later experiments with the parameters $\alpha=5$ and $\beta=5$. All in all, The recall close to 90\% when alpha and beta equal 5 may provide evidence for our two assumptions mentioned above.

\subsection{Quality of Candidate List}

\begin{figure}
\includegraphics[width=\textwidth]{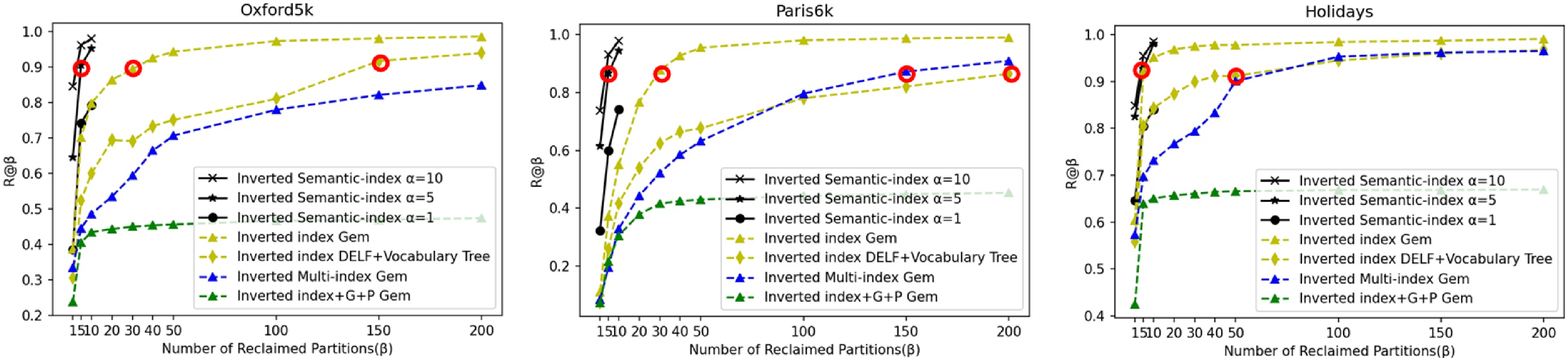}
\caption{The recall performance of several index with different features are shown here. $R@\beta$ represents the recall rate when $\beta$ partitions are reclaimed from the total $N_l=1k$ blocks. With our inverted semantic-index, the similar points are gathered more closely and a candidate list with higher quality is produced.} \label{fig2}
\end{figure}

It is proved that the recall in semantic-index is about 90\% with $\alpha=5$ and $\beta=5$ in the last section. Now we compare the quality of the generated candidates with the state-of-the-art indexes when the number of selected partitions  $\beta$ is set to 1, 5, 10, 20, 30, 40, 50, 100, 150 and 200 (In Fig.~\ref{fig2}, we can see that the recall is nearly 100\% with our Inverted Semantic-index at $\beta=10$, thus the experiments with $\beta>10$ are not performed).
As can be seen in Fig.~\ref{fig2}, the  most suitable number of reclaimed partitions $\beta$ is normalized to 5 or 10 in our index, while this parameter can not be determined without the knowledge of the data distribution in other indexes. For example, in order to reach the recall of semantic-index at top5-5, the corresponding $\beta$ of oxford5k, Paris6k and Holidays are $(30,30,5)$ in the classical IVF with Gem feature, $(150,200,50)$ in the IVF with DELF+vocabulary-tree feature and $(>200,150,50)$ in Inverted Multi-index (IMI) with Gem. In other words , the most suitable number of reclaimed partitions $\beta$ relys on the data distribution and features type. Nevertheless, the setting of parameter $\beta$ in our inverted semantic-index is robust no matter what is the dataset or feature extractor, which may indicate that the data arrangement is more reasonable in our inverted semantic-index. 
\subsection{Experiment on Different Features}
 In order to evaluate the retrieval accuracy of our index on the real world image retrieval datasets, we carry out experiments with two state-of-the-art  feature extractors DELF~\cite{delf} and Gem~\cite{Gem}.
 
\setlength{\tabcolsep}{8pt}
\begin{table}
\begin{center}
\caption{Experiments with local feature DELF }
\label{tab1}
\begin{tabular}{lllllll}
\hline
Index & $mAP$ & $Recall$&$R@1$&$R@10$&$R@100$&$t(s)$\\
\hline
\multicolumn{7}{c}{\bf Oxford5k, $ \alpha=5, \beta=5$}\\
\hline
Exhaustive    & 0.848  & 1     &0.059 &	0.416 &	0.881 &3193.459\\
 Ours           & 0.792 & 0.939 &0.059  &0.408   &0.799   &2896.929\\
\hline
\multicolumn{7}{c}{\bf Paris6k, $ \alpha=5, \beta=5$}\\
\hline 
Exhaustive    & 0.826   & 1     &0.008 &	0.078 &	0.564   &4049.376\\
 Ours           & 0.746  & 0.866  &0.008 &	0.078 &	0.557   &2605.226 \\
 \hline
\multicolumn{7}{c}{\bf Holidays, $ \alpha=5, \beta=5$}\\
\hline 
Exhaustive    &0.833    & 1     &0.393 &	0.788 &	0.819  &1343.571\\
 Ours           & \bf0.838   &0.927  &0.393 &\bf0.848 &	\bf0.916   & 359.775 \\
\hline
 \hline
\multicolumn{7}{c}{\bf Oxford105k, $ \alpha=5, \beta=5$}\\
\hline 
Exhaustive    &-    & -     &- &	- &	-   &-\\
 Ours           & 0.806  &0.939   &0.059 &	0.409 &0.819    & 13784.351 \\
\hline
\end{tabular}
\end{center}
\end{table}
\setlength{\tabcolsep}{1.4pt}
\subsubsection{Local Feature DELF}First we use the outstanding local feature DELF~\cite{delf} in our retrieval system. Thanks to the geometric processing of RANSAC, DELF has achieved excellent accuracy in instance image retrieval. However the speed of DELF is too slow for large-scale retrieval, thus the acceleration is necessary. As a result of the unfixed feature dimension and the complex comparison methods based on RANSAC, the incompatibility of DELF to current cluster-based index occurs. On the contrary, the construction of our inverted semantic-index does not rely on the features in the database. Thus it is feasible to combine DELF with our Semantic-index so as to improve the retrieval efficiency.

Compared to exhaustive search, the results in Tab.~\ref{tab1} reveal that with our inverted semantic-index, the accuracy of retrieval $mAP$ is remained with significant speed improvement, which is even improved in Holidays. The improvement in Holidays is attributed to the high candidate list quality that eliminate the interference of some irrelevant samples.  Although the speed is still too slow due to the complex RANSAC comparison in DELF, our inverted semantic-index still work with local features at least unlike other indexes.
\setlength{\tabcolsep}{6pt}
\begin{table}
\begin{center}
\caption{Experiments with global feature Gem}
\label{tab2}
\begin{tabular}{llllllllllllll}
\hline
Index &$N_l$&$\beta$& $mAP$ & $Recall$&$R@1$&$R@10$&$R@100$&$t(s)$\\
\hline
\multicolumn{9}{c}{\bf Oxford5k, $ \alpha=5$}\\
\hline
IVF+G+P~\cite{g+p}&1k&5& 0.393  & 0.405  &0.059 &	0.307 &	0.405    &1.400
\\
IVF~\cite{ivf}&1k&5& 0.650   &0.701   &0.059 &\bf0.425& 	0.701&1.396 \\

Multi-index~\cite{imi}&$2^{10}$&5& 0.440  & 0.445  &0.059 &	0.355 &	0.445  &9.580 \\

Semantic-index&1k&5& \bf0.778  &\bf0.939  &0.059 &	0.407 &	\bf0.793  &3.183 \\
\hline
IVF-ADC~\cite{pq}&1k&5& 0.630  & 0.701  &0.059 &\bf0.425& 	0.701   &1.407 \\

Multi-ADC~\cite{imi}&$2^{10}$&5& 0.432  & 0.445  &0.059 &	0.355 &	0.445  
 &11.975 \\
Semantic-ADC&1k&5&\bf0.747  & \bf0.939  &0.055 &	0.373 &\bf0.788   &3.849\\
\hline
\multicolumn{9}{c}{\bf Holidays, $ \alpha=5$}\\
\hline
IVF+G+P~\cite{g+p}&1k&5& 0.636   & 0.639   &0.392 &	0.639 &	0.639  &1.590
  \\
IVF~\cite{ivf}&1k&5& \bf0.903    &\bf0.921    &0.392 &	\bf0.920 &	\bf0.921 &1.822  \\

Multi-index~\cite{imi}&$2^{10}$&5& 0.695   & 0.697   &0.392 &	0.697  & 0.001 
 &11.578  \\

Semantic-index&1k&5& \underline{0.874}  & \underline{0.916}   &0.392 &	\underline{0.889} &	\underline{0.915}     &2.942  \\

\hline
IVF-ADC~\cite{pq}&1k&5& \bf0.896   & \bf0.921   &0.392 &	\bf0.920 &	\bf0.921   &1.823   \\
Multi-ADC~\cite{imi}&$2^{10}$&5&0.695  & 0.697   &0.392 &0.697 	&0.697  &12.675  \\
Semantic-ADC&1k&5&\underline{0.779 }    & \underline{0.915}      &0.337 &	\underline{0.869} &	\underline{0.911}   &2.970
\\
\hline
\end{tabular}
\end{center}
\end{table}
\setlength{\tabcolsep}{1.4pt}
\subsubsection{Global Feature Gem} In order to compare with the cluster-based state-of-the-art index~\cite{imi}~\cite{g+p}, we carry out experiments with a state-of-the-art global feature Gem~\cite{Gem} and the results are shown in Tab.~\ref{tab2}. As can be seen, the performance of IMI and IVF+G+P is not satisfactory when the database scale is small compared with our index. On the harder dataset Oxford, our inverted semantic-index performs best with the accuracy $77.8\%$, while  the precision of the semantic-index is close to that of the classical IVF on the easier dataset Holiday. In conclusion, the performance of the inverted semantic-index is competitive with the state-of-the-art methods on the small-scale datasets.  

\subsection{Merging and Splitting the partitions}

For the purpose of making the codebook size changeable, we propose the merging and splitting method in the section 3.2. In this section, we will show the effectiveness of the method by experiments. For codebook reduction, the $1k$ partitions are merged into 581 cells based on the label similarities.  For codebook enlargement, we build 10 sub-codebooks under each semantic partition by K-means such that a codebook of size $100k$ is generated. Experiments on Oxford5k are complied and shown in Tab.~\ref{tab4}, where the ratio of scope $r_s$ is the percentage of the candidate list compared with the whole database. 
\setlength{\tabcolsep}{8pt}
\begin{table}
\begin{center}
\caption{Experiments of merged or split semantic partitions on Oxford5k with Gem}
\label{tab4}
\begin{tabular}{llllllllllllll}
\hline
$N_l$&$\beta$& $mAP$ & $Recall$&$R@1$&$R@10$&$R@100$&$r_s$&$t(s)$\\
\hline
581&5& 0.787  & 0.950  &0.059 &	0.411 &	0.802  &\underline{0.525}  &3.630  \\
1k&5& 0.778  &0.939  &0.059 &	0.407 &	0.793    & \underline{0.469} &3.183
 \\
10k&45& 0.771  &  0.939   & 0.055 &	0.405 &	0.793    & \underline{0.456}   & 3.504
  \\
10k&55&  0.766   & 0.944   & 0.049 &	0.408 &	0.796  &  \underline{0.478}   & 3.485
  \\
\hline

\end{tabular}
\end{center}
\end{table}
\setlength{\tabcolsep}{1.4pt}

As shown in Tab.~\ref{tab4}, each cell includes more points with a smaller codebook, thus the recall is higher and the retrieval time is longer. On contrast, the database is divided more finely with a larger codebook such that the adjustment of searching scope $\beta$ is more flexible. For example, We can slightly narrow the searching scope from 0.469 to 0.456 by reduce $\beta$ from 50 to 45 with no decline of recall (The result of 5 from $1k$ is the same as that of 50 from $100k$).

\subsection{Experiment on Large-Scale Datasets}
The datasets Oxford5k, Paris6k and Holidays presented previously are too small. In order to test out Inverted Semantic-index on large-scale datasets, we carry out the experiments on Oxford105k and Oxford1M which consist of Oxford5k and Flickr100k, Flikr1M.
\setlength{\tabcolsep}{3pt}
\begin{table}
\begin{center}
\caption{Experments on large-scale datasets with feature Gem}
\label{tab3}
\begin{tabular}{llllllllllllllll}
\hline
Index &$N_l$&$\beta$& $L$&$\tau$& $mAP$ & $Recall$&$R@1$&$R@10$&$R@100$&$t(s)$\\
\hline
\multicolumn{11}{c}{\bf Oxford105k, $ \alpha=5$}\\
\hline

IVF+G+P~\cite{g+p}&1k&5&64&0.5& 0.535 &0.581 &0.059 &	0.369 &	0.566   & 1.749  \\
IVF~\cite{ivf}&1k&5&-&-&  \underline{0.744}  &0.874  &0.059 &	0.385 &	0.761  & 2.671     \\

Multi-index~\cite{imi}&$2^{10}$&5&-&-& 0.549 & 0.574   &0.059 &	0.340 &	0.566  &21.547 \\

Semantic-index&1k&5-&-&& \bf0.756   & \bf0.939    &\bf0.059 &\bf	0.406 &	\bf0.776     & 6.769  \\
\hline
ADC~\cite{pq}&1&1&-&-& 0.545  &0.989  &0.054 &	0.256 &	0.646   & 123.757   \\
IVF-ADC~\cite{pq}&1k&5&-&-& \underline{0.485}     & 0.874    &0.027 &	0.159 &	0.635  &4.777   \\
Multi-ADC~\cite{imi}&$2^{10}$&5&-&-&  0.507 & 0.574 &  \bf0.059 	&0.299& 0.559  &  19.513  \\
Semantic-ADC&1k&5&-&-& \bf0.702  &\bf0.939 &	0.041 &	\bf0.340 &\bf	0.790  &12.702 \\
\hline
\multicolumn{11}{c}{\bf Oxford1M, $ \alpha=5$}\\
\hline

IVF+G+P~\cite{g+p} &1k&5&64&0.5& 0.627    & 0.726    &0.059 &	0.395 &	0.673   &32.287   \\
IVF~\cite{ivf}&1k&5&-&-& \underline{0.676}  &0.803  &0.059 &	0.385 &	0.689   &114.284  \\
Multi-index~\cite{imi}&$2^{10}$&5&-&-& 0.482&0.533& 0.059 &0.291 &0.505 &252.228  \\
Semantic-index&1k&5&10&0.1& 0.599  & 0.915    & 0.003 &	0.375 &	0.729  &  167.751  \\
Semantic-index&1k&5&-&-& \bf0.719   & \bf 0.939  & \bf 0.059& \bf	0.399 &	\bf0.735 &  1272.978  \\
\hline
ADC~\cite{pq} &1&1& -  & - &0.133 &	 0.798 & 0.030&	0.078&	0.198&8202.504

    \\
IVF-ADC~\cite{pq}&1k&5&-&-& \underline{0.102 }   & 0.803    &0.004 &	0.013 &	0.161     &56.894   \\
Multi-ADC~\cite{imi}&$2^{10}$&5&-&-& 0.399  &  0.533  & \bf0.059 &	\bf0.239 &	0.445   & 245.490 \\
Semantic-ADC&1k&5&-&-&\bf0.493 &\bf0.939 &0.028 &	0.178 &\bf0.623  &1031.553 
 \\
\hline
\end{tabular}
\end{center}
\end{table}
\setlength{\tabcolsep}{1.4pt}

As we can see, the application of PQ only causes little drop in accuracy when the dataset scale is very small in Tab.\ref{tab2}. On the contrary, when we enlarge the data scale to $100k$ and $1M$, the accuracy loss of IVF caused by PQ compression becomes more serious as shown in Tab.~\ref{tab4}, where the drops are up to 25.9\% and 57.4\%. We believe the reason might be the unreasonable cluster-based division on the top level. With our inverted semantic-index, the accuracy decline is more acceptable at 5.4\% in Oxford105k and 22.5\% in Oxford1M. Thanks to the participation of semantic information, only the images related semantically are gathered in one partition. Therefore, the detailed distinction between points are learned by PQ in our index. 

In order to prove that the boundaries on the top level can influence the PQ encoding quality, we perform ablation experiments (ADC) by ranking compressed data without inverted indices and reclaiming the first $20k$ data (Re-ranking all the points takes too much time). The accuracy of ADC is 54.5\%  on $100k$ scale, which is higher than 48.5\% and lower than our 70.2\%. Consequently, our inverted-index can not only remain the retrieval performance but also help PQ to improve the precision from 54.5\% to 70.2\% on the 100k scale and from 13.3\% to 49.3\% on the 1 million scale. On the contrary, the accuracy in IVF-ADC is lower than pure ADC, despite that the accuracy and recall of IVF without PQ is competitive with ours. To sum up, our inverted semantic-index can not only remain the accuracy of image retrieval on all-scale datasets, but also improve the representation ability of PQ. 

As for the IVF+G+P~\cite{g+p} and IMI~\cite{imi}, their performance is not very satisfactory on the small-scale datasets despite improvements on the large-scale dataset, since they are designed for large-scale. Otherwise, our inverted semantic-index achieves better accuracy robustly regardless of the database size. However, our retrieval efficiency is not very high compared with others, which is due to the implement based on python in a single thread mode, while the implements in Faiss are usually in several threads. Another reason is that there are too many points under each semantic partition, thus we enlarge the codebook size into 10k and pruning the sub-cells in each semantic partition using the method mentioned at the end of 3.2. When we only remain 10\% of the data, the recall is 91.5\% but the time decreases to 167.751 seconds, which proves the effectiveness of the splitting and pruning operations in the semantic partition.

\section{Conclusions}
In this paper, we propose a new semantic-index for large-scale instance image retrieval. The index can remain the retrieval accuracy and reduce the precision loss caused by the PQ quantization. The experiments prove our inverted semantic-index achieves outstanding performance compared with the state-of-the -art works. We will make efforts to improve the speed performance and attempt to combine instance retrieval with content-based retrieval based on the exclusive Inverted Semantic-index in the future.    

\bibliographystyle{unsrt}  
\bibliography{eccv}

\end{document}